\documentclass[final]{cvpr}

\usepackage{times}
\usepackage{epsfig}
\usepackage{graphicx}
\usepackage{amsmath}
\usepackage{amssymb}
\usepackage{cuted} 
\usepackage[font={small}]{caption}
\usepackage[x11names]{xcolor}
\usepackage{verbatim}
\usepackage[numbers,sort,compress]{natbib}
\usepackage{booktabs}
\usepackage{nicefrac}

\DeclareMathSymbol{\jj}{\mathalpha}{letters}{"11}
\newcommand{\bTheta}{\Theta}
\newcommand{\btheta}{\boldsymbol{\theta}}
\newcommand{\bbeta}{\boldsymbol{\beta}}

\newcommand{\bphi}{\boldsymbol{\phi}}

\definecolor{mypink1}{rgb}{0.858, 0.188, 0.478}

\DeclareMathSymbol{@}{\mathord}{letters}{"3B}

\newcommand{\denselist}{\itemsep 0pt\parsep=0pt\partopsep 0pt\vspace{-\topsep}}

\usepackage[pagebackref=true,breaklinks=true,colorlinks,bookmarks=false]{hyperref}

\usepackage{cleveref}

\begin{document}

\title{D3D-HOI: Dynamic 3D Human-Object Interactions from Videos}
\author{
Xiang Xu$^{1}$\thanks{Work done when interning at Facebook AI Research.} \quad Hanbyul Joo$^2$ \quad Greg Mori$^1$ \quad Manolis Savva$^1$\\[3pt]
$^1$Simon Fraser University \quad $^2$Facebook AI Research\\
}
\maketitle

\begin{strip}

\includegraphics[width=\linewidth]{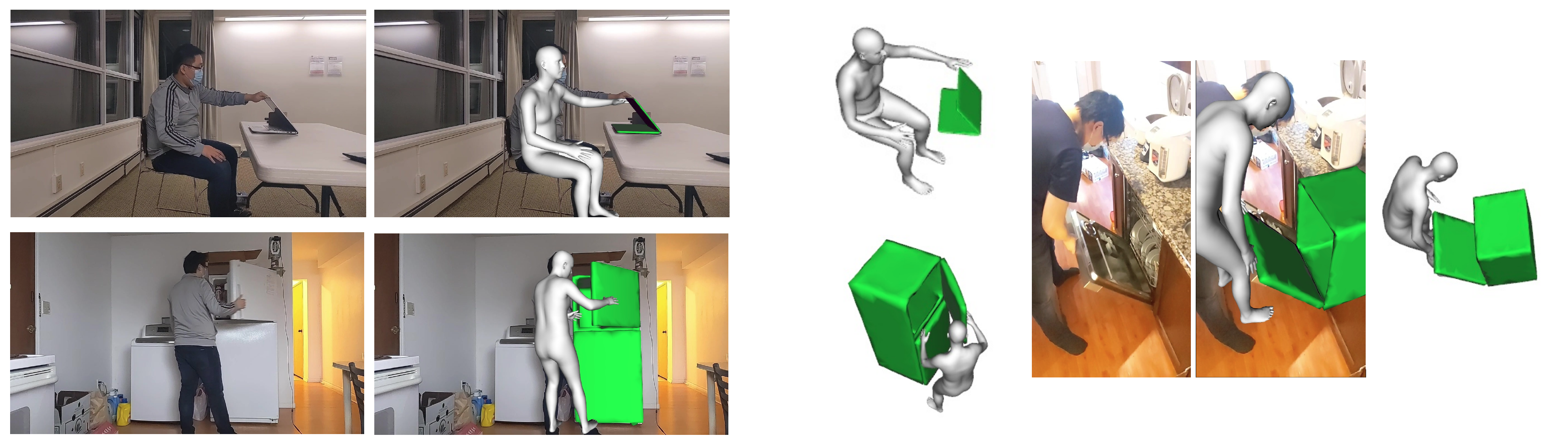}
\captionof{figure}{
We address the problem of recovering dynamic 3D human-object interactions given a video input.
Our focus is on reconstructing the articulating 3D object during manipulation. To enable research in this direction, we collect a dataset of interaction videos with common objects (e.g., laptop, fridge, dishwasher). We annotate the 3D object pose, shape, articulation state, and estimate the 3D mesh of the manipulator using the approach from \citet{joo2020exemplar}. Above are rendered frames from the ground truth annotations.
}
\label{fig:teaser}
\end{strip}

\begin{abstract}

We introduce D3D-HOI: a dataset of monocular videos with ground truth annotations of 3D object pose, shape and part motion during human-object interactions.
Our dataset consists of several common articulated objects captured from diverse real-world scenes and camera viewpoints. 
Each manipulated object (e.g., microwave oven) is represented with a matching 3D parametric model.
This data allows us to evaluate the reconstruction quality of articulated objects and establish a benchmark for this challenging task.
In particular, we leverage the estimated 3D human pose for more accurate inference of the object spatial layout and dynamics.
We evaluate this approach on our dataset, demonstrating that human-object relations can significantly reduce the ambiguity of articulated object reconstructions from challenging real-world videos. Code and dataset are available at \href{https://github.com/facebookresearch/d3d-hoi}{https://github.com/facebookresearch/d3d-hoi}.

\end{abstract}

\section{Introduction}

3D reconstruction has gained prominence in recent years. The goal is to infer the full 3D shape, pose and layout from only partial 2D information. This is challenging due to the inherent ambiguities caused by mapping back from 2D to 3D.
Most works in this area focus on separately reconstructing the human mesh~\cite{kanazawa2018end,bogo2016keep,kolotouros2019learning,pavlakos2019expressive,joo2020exemplar,kocabas2019vibe} or the individual 3D object~\cite{tulsiani2018factoring,gkioxari2019mesh, maninis2020vid2cad, li2020category}. Recent works~\cite{savva2016pigraphs, hassan2019resolving, zhang2020perceiving, weng2021holistic} do explore human-object relations for more accurate 3D reconstruction, but they only deals with static objects with no part motion.

Real world human-object interactions (HOIs) often involves dynamic articulated objects (e.g., opening and closing a fridge). Traditional approaches~\cite{li2020category, xiang2020sapien, huang2021multibodysync} for articulated object reconstruction requires accurate RGB-D data and do not leverage the human-object relations. In response to the lack of 3D HOI data, we first create the Dynamic 3D HOI (D3D-HOI) dataset. We then provide an optimization-based method that utilize human-object relations to reconstruct the articulated objects from only RGB video. 

D3D-HOI is a real world HOI video dataset with 3D annotations of object location, orientation, size, matching CAD model, and part motion state at every video frame.
Our annotations are based on 3D mesh models from \citet{xiang2020sapien} which we use to represent objects with moving parts that articulate along a specific motion axis and origin.
We focus on translation (prismatic joint) and rotation (revolute joint) motions. For simplicity, we only consider one joint motion per object part. Note that an object can have multiple parts and each part will have its own unique joint motion. 
\Cref{fig:teaser} illustrates rendered examples of the ground truth annotations.
D3D-HOI allows us to evaluate reconstruction methods along several axes: shape, pose and part motion of real-world articulated objects during human-object interaction.

Using the D3D-HOI dataset, we explore the role of human-object relations and propose an optimization method that leverages the estimated human pose and dynamics to better reconstruct the interacted object.
Our insight is that treating both the human and object as dynamic entities allow us to constrain the pose and motion of each through orientation and contact terms.
To the best of our knowledge, we are the first to reconstruct articulated object from real-world HOI video.

In summary, we make the following two contributions:
\begin{itemize}\denselist
  \item We collect a video dataset of human interactions with articulated objects and provide the ground truth annotations of object 3D pose, shape and part motion at every time frame.
  \item We present an optimization method based on human-object relations for the task of articulated object reconstruction from HOI video. We also evaluate the 3D reconstruction accuracy on our new dataset.
\end{itemize}

\section{Related Work}

\subsection{3D human pose and motion estimation}
Recently there has been a lot of interest for estimating human pose and motion from 2D image or video. Fitting-based methods assume a parametric body model such as SMPL~\cite{loper2015smpl} or SMPL-X~\cite{pavlakos2019expressive} and use optimization algorithms like SMPLify~\cite{bogo2016keep} to fit the parameters. On the
other hand, regression-based methods~\cite{kanazawa2018end,Omran2018NeuralBF,Pavlakos2018LearningTE} rely on deep neural networks and large amounts of training data~\cite{h36m_pami, Marcard2018RecoveringA3} to directly predict the 3D pose of the human. Hybrid methods~\cite{kolotouros2019learning, joo2020exemplar} also demonstrate improved results. Compared to image-based methods, video-based methods~\cite{kocabas2019vibe, Luo_2020_ACCV} can utilize temporal information and better estimate human pose and motion over entire video sequences. 

\subsection{3D object pose and motion estimation}
Recovering object shape and pose from 2D images is a very active field. Recent methods~\cite{tulsiani2018factoring,kulkarni20193d,gkioxari2019mesh,nie2020total3dunderstanding} can achieve impressive results on indoor environments. These methods are designed to work on non-articulated objects where object state is fixed and there is no part motion. 

Methods that do predict part motion parameters usually come from the field of embodied ai, where the goal is to teach robots how to open or close doors. Synthetic datasets~\cite{xiang2020sapien,shen2020igibson} are used to train regression-based models to directly predict part motion parameters. However, methods train on these dataset require accurate depth information~\cite{li2020category,jain2020screwnet, huang2021multibodysync, yan2020rpm}. Thus it is not straightforward for one to directly apply them to real world scenes due to noisy depth estimation and domain gap differences. Also none of these methods utilize human information during HOI. 

\subsection{3D human-object interaction}
Some recent works do examine 3D human-object relations during HOI and use them to improve performance on various tasks. PiGraphs~\cite{savva2016pigraphs} collects a dataset with 3D annotation of human and interacted objects (i.e. chair, desk, whiteboard). Hassan et al.~\cite{hassan2019resolving} introduced the PROX dataset which consists of 3D human mesh and static 3D scene. They use static 3D scene structure to improve human pose estimation from monocular images. Follow-up works~\cite{PLACE:3DV:2020,hassan2020populating,zhang2020generating} also use this dataset to generate human mesh and insert it to an environment based on 3D scene context. Human-object relations are also important for 3D reconstruction. In~\cite{Chen_2019_ICCV}, the authors build a graph based on human-object relations and propose a holistic scene parsing and reconstruction pipeline. Zhang~\cite{zhang2020perceiving} jointly infers spatial layout and shapes of humans and objects in a consistent 3D scene. Karunratanakul~\cite{karunratanakul2020grasping} defines a grasping field to jointly reconstructs the human hand and interacted object. For those works, only the human has motion and the object remains static. This is different from our dynamic human-object interaction setting where both object and human can have motions.

\vspace{20pt}
\begin{table*}[]
\resizebox{\textwidth}{!}{%
\begin{tabular}{@{}lcccccccc@{}}
\toprule
          & refrigerator & storageFurniture & trashcan & washingmachine & microwave & dishwasher & laptop & oven \\ \midrule
video     & 82           & 32               & 19       & 24             & 36        & 11         & 41     & 11   \\
cad model & 7            & 4                & 2        & 3              & 2         & 2          & 2      & 2    \\
viewpoint & 14           & 5                & 4        & 7              & 7         & 4          & 8      & 3    \\
scene     & 5            & 4                & 3        & 3              & 2         & 2          & 2      & 1    \\ \bottomrule
\end{tabular}
}
\caption{\label{tab:dataset} Statistics of the Dynamic 3D HOI Dataset with category-level distributions for number of videos, 3D CAD models, camera viewpoints, and 3D scenes.}
\end{table*}

\section{Dynamic 3D HOI Reconstruction\label{sec:pro_state}}
Given a video from a fixed camera viewpoint where a human dynamically interacts with an articulated object, we aim at estimating the 3D human motion, 3D movement of the articulated object, and their 3D arrangement. In particular, we use articulated parametric models to represent humans and objects respectively. This takes into 
account the dynamic interactions between them while ignoring detailed shape variations.  
Specifically, we use the SMPL model~\cite{loper2015smpl} to represent the 3D state of a human at time $t$ by $\bTheta_{h}^t = ( \boldsymbol{x}_h^t, \btheta_{h}^t, \bbeta_{h})$, where $\boldsymbol{x}_h^t \in \mathbb{R}^3$ is 3D translation,  $\btheta_{h}^t \in \mathbb{R}^{24\times 3}$ controls the rotations of 24 body joints with respect to their parent joints, and the shape $\beta_{h} \in \mathbb{R}^{10}$ controls the shape of the body. Similarly, we represent each 3D object at time $t$ via a parametric model $\bTheta_{o}^t = ( \boldsymbol{x}_o, \bphi_{o}, \btheta_{o}^t, \beta_{o})$, where $\boldsymbol{x}_o \in \mathbb{R}^3$ is the global translation and $\bphi_{o} \in \mathbb{R}^{3\times3}$ is the global orientation of the object base. In particular, $\btheta_{o}^t \in \mathbb{R}^n$ represents the 3D part motion value of the dynamic part. This is either rotation degree for revolute joint or translation offset for prismatic joint. The object shape parameter $\bbeta_{o} \in \mathbb{R}^{3}$ represents the width, length, and height sizes of the object. Thus, our target task is to estimate $\{ \bTheta_{h}^t\}_{t=[1,n]}$ and  $\{ \bTheta_{o}^t\}_{t=[1,n]}$ from a monocular video, $\{I_t\}_{t=[1,n]}$.  Note that $\bTheta_{h}^t$ and $\bTheta_{o}^t$ are defined in the same 3D world space. Examples are shown in Fig~\ref{fig:teaser}. Since there has been a lot of works for 3D human reconstruction using SMPL, we use an existing method~\cite{joo2020exemplar} to extract the 3D human shape and pose. The focus of this work is thus on the estimating $\{ \bTheta_{o}^t\}_{t=[1,n]}$.

\begin{figure*}[t]
\includegraphics[width=1.0\linewidth]{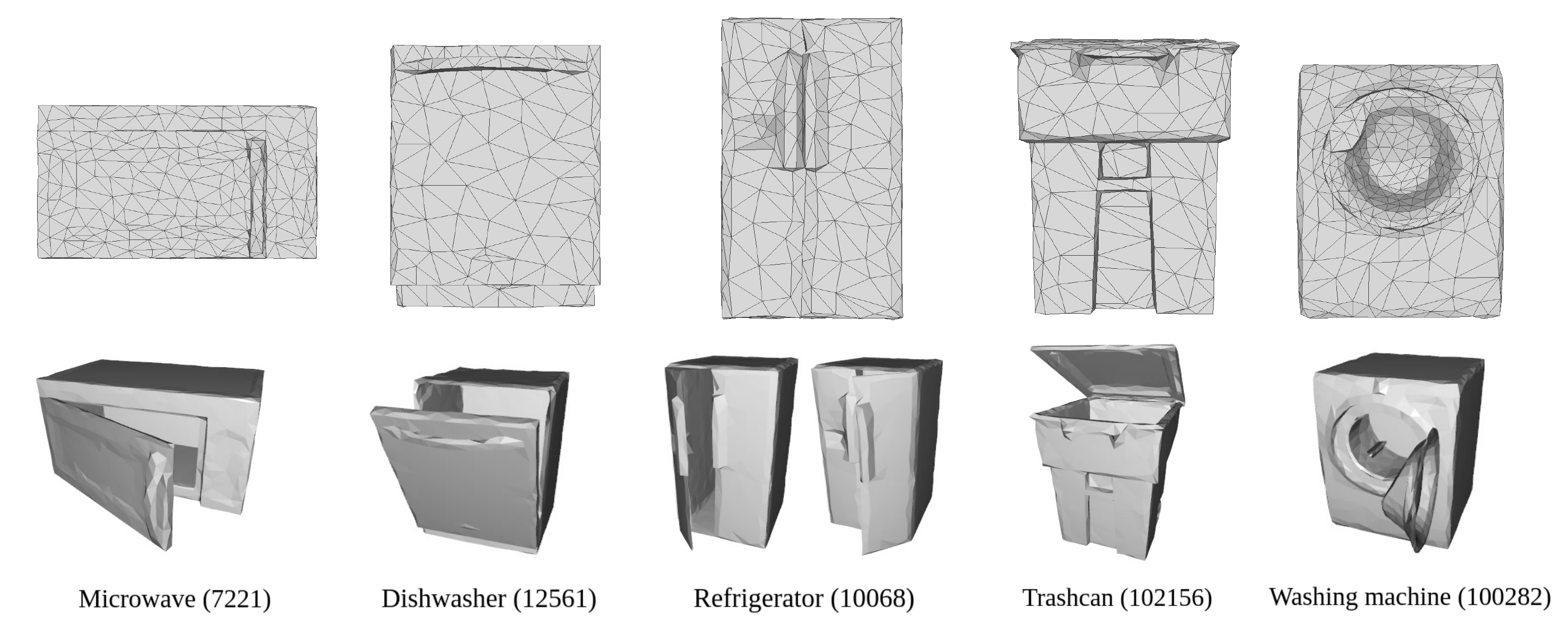}
\captionof{figure}{Examples of post-processed meshes from the PartNet-Mobility dataset~\cite{xiang2020sapien}. Top row shows the simplified meshes with centering and front direction in $+z$ direction. Bottom row illustrates the available part motions for each object.}
\label{fig:cad}
\end{figure*}

\section{Dynamic 3D HOI Dataset}
A major obstacle in pursuing the study of dynamic 3D HOI reconstruction task is the lack of available datasets on which algorithms are quantitatively compared and evaluated. To tackle this challenge, we create the Dynamic 3D HOI Dataset (D3D-HOI), providing videos of dynamically interacting individuals\footnote{The videos have been captured by the authors, with approvals from the individuals appearing in the videos to release the data for research purposes.}, with manually annotated ground-truth 3D object parameters, $\{ \bTheta_{o}^t\}_{t=[1,n]}$. In this section, we describe the statistics and summary of our dataset (subsection \ref{sec:data_stats}), the data processing procedures to build 3D articulated models (subsection \ref{sec:data_preprocess}), and the data collection pipeline for video capture and annotations (subsection \ref{sec:data_collect}),

\subsection{Data Statistics \label{sec:data_stats}}
D3D-HOI contains a total of 256 videos. After subsampling videos into 3 FPS, our dataset has 6286 image frames in total. Each frame is annotated with 3D object dimension, matching CAD model, location, orientation, and part motion. Data acquisition involved 5 volunteers interacting with objects in 22 different scenes. The articulated objects come from 8 SAPIEN categories and were represented by a total of 24 CAD models. The camera is placed at 52 different viewpoints from the object. Table~\ref{tab:dataset} summarizes the overall dataset statistics. From the summary we see that the collected dataset contains a variety of articulated objects from small (e.g., laptop, microwave) to large (e.g., refrigerator, dishwasher). The distribution over viewpoints, scenes, and 3D models are also very diverse and covers a wide range of human-object interaction scenarios. Example annotated frames from our dataset is shown in Figure~\ref{fig:teaser}.

\subsection{Articulated Models \label{sec:data_preprocess}}
We consider object categories that are commonly observable in our daily life and have movable parts with which humans can interact: refrigerator, storage furniture (including drawer), microwave, dishwasher, refrigerator, trash can, washing machine, laptop, and oven.
To build the parametric models for articulated objects, we leverage the SAPIEN PartNet-Mobility~\cite{xiang2020sapien} dataset by applying additional post-processing to simplify mesh topology. The PartNet-Mobility dataset provides various 3D articulated mesh models with movable part definitions. Among the models in our target object categories, we choose the 3D
CAD models (up to four per category) most similar to the objects captured in our input videos. We also rescale the 3D models to have the same size with the actual objects we captured in measurement units (height, width, and length in $cm$). We define a local object coordinate frame by positioning the origin at the center of mesh and by making the ``frontal'' side of the object to be aligned with the $+z$ direction in the local coordinate. The $+y$ direction is defined as the ground normal direction when object is normally sitting on the floor. Finally, we simplify mesh topology to contain roughly 2500 triangles per object. Figure~\ref{fig:cad} shows examples of processed PartNet-Mobility models and their available part motions.

\begin{figure}[t]
\includegraphics[width=0.5\textwidth]{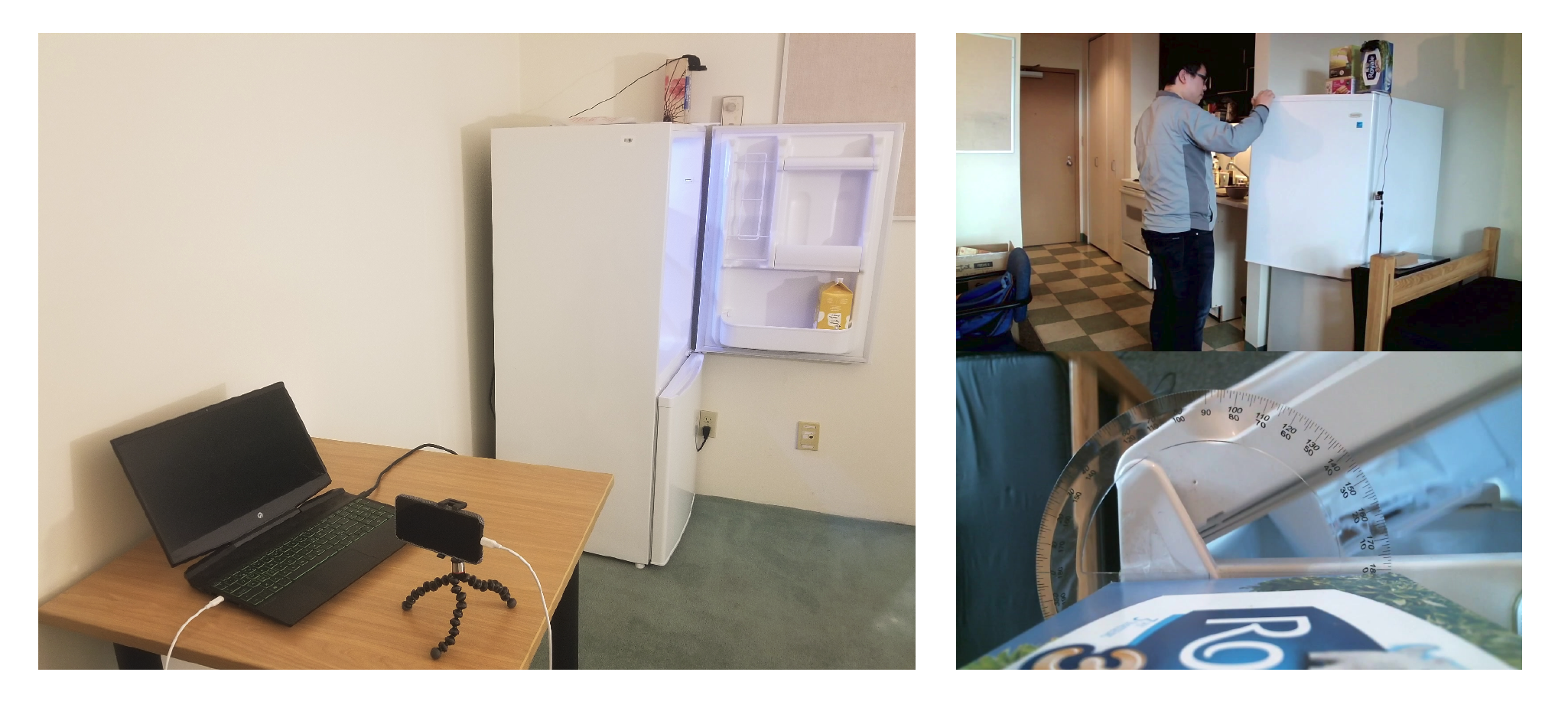}
\caption{Left is an example of our equipment setup with dual cameras (one for recording, one for reading motion parameters). Right shows an example of the synchronized raw frame.}
\label{fig:equipment}
\end{figure}

\subsection{Data Collection Pipeline \label{sec:data_collect}}

\paragraph{Capture System} We use two cameras (Google Pixel 4 and AVerMedia live streamer 313) to simultaneously record RGB videos on the HOI scenes from two different view points. One camera (Pixel 4) records the scenes from a fixed third person view where the whole human body is visible, and another camera (AVerMedia) is placed on the object to precisely record the movement of object part with a measurement tools (e.g., protractor). See Figure~\ref{fig:equipment}. This multi-camera setup is helpful to manually annotate the parameters of movable parts (i.e. angles in degrees) at each video frame. All videos are recorded at 30 FPS.

\paragraph{Calibration} To estimate the intrinsic camera parameters, we perform camera calibration for each camera we used with a checkerboard pattern using OpenCV~\cite{opencv_library}. 
Since the front camera of Pixel 4 and the AVerMedia live streamer 313 have fixed focus, the calibration parameters are valid for all captured videos. 
However, for the rear Pixel 4 camera with auto-focus, our calibration parameters estimated afterwards may not be precise. To check this issue, we perform camera calibration multiple times by locking the focus with various target object distances similar to our capture setup (1.5 to 3.0 meters away from the camera). We compared the parameters, and confirmed that after locking auto-focus has minor intrinsic parameter differences. To this end, we use the averaged focal length estimated at different distance as the intrinsic parameters for this camera.

\begin{figure}[t]
\includegraphics[width=0.5\textwidth]{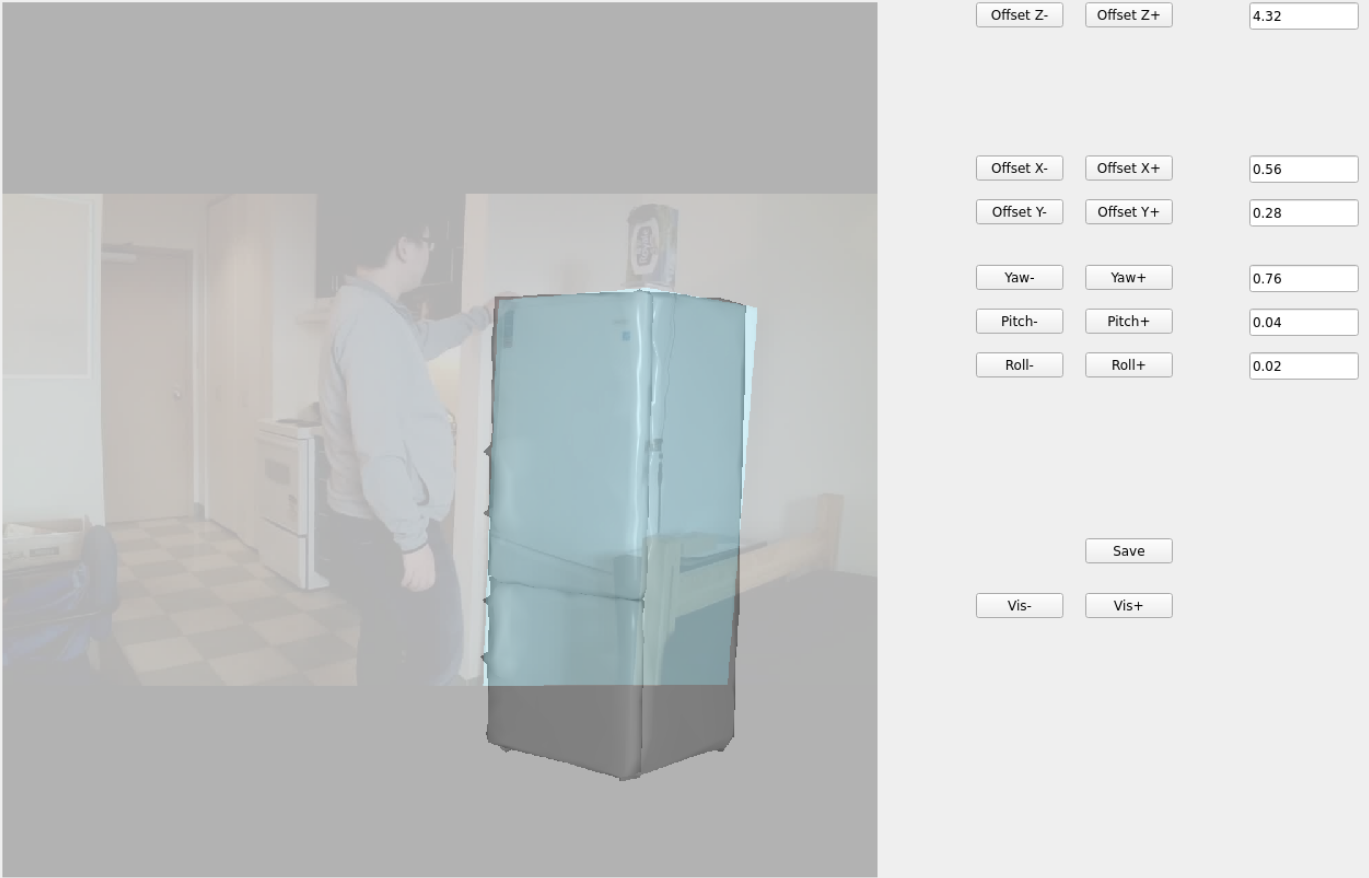}
\caption{Interface for annotating the object location and orientation values.}
\label{fig:interface}
\end{figure}

\paragraph{Annotation Interface} 
We build an interactive GUI interface for annotating the 3D orientation and location of the object. A screenshot of the interface is shown in Figure~\ref{fig:interface}. In our annotation interface, users can see the input image frames together with the projection of 3D objects using the current object parameters. For the projection, the calibrated intrinsic parameters are used. Users can manually adjust the object parameters: translation parameter $\boldsymbol{x}_o$ and orientation parameter $\bphi_{o}$. The 3D object is re-rendered with the adjusted parameters in real-time so that users can confirm that the object projection is aligned to the input image. We save the final values as ground-truth annotation data. Note that these translation and orientation parameters are fixed across frames, since the videos are recorded from a fixed camera view point and objects are static except the movable parts. The parameters for the movable part (e.g. doors) are separately annotated by observing the videos recorded with measurement tool (e.g. protractor) and annotating the angles at 3 FPS.

\section{Methods}
We first describe our design of the differentiable articulated object in~\ref{sec:diff_part}. We then go over the objective function in Section~\ref{sec:final_loss}. Section~\ref{sec:obj_only} provides detailed description of the object mask error. Section~\ref{sec:h_fit} shows how we fit human mesh to the same space as the object. Human-object interaction error is discussed in Section~\ref{sec:obj_pp_joint}. Details of object parameter regularization is in Section~\ref{sec:constraint}. And finally we discuss the optimization pipeline in Section~\ref{sec:optim}.

\section{Differentiable Articulated Object \label{sec:diff_part}}
In this section, we describe how we differentiate from 3D object mesh to the articulated object parameters. Let us denote the processed object mesh $\mathbf{V} = \{\mathbf{v}_i\}$ where $\mathbf{v}_i \in \mathbb{R}^3$ is the $i$-th vertex location at the local object coordinate (local coordinate with origin at the center). The object mesh can be divided into movable part and non-movable part (base): $\mathbf{V} = \mathbf{V}_{\text{part}} \cup \mathbf{V}_{\text{base}} $. Given object parameters $\bTheta_{o}^t = ( \boldsymbol{x}_o, \bphi_{o}, \btheta_{o}^t, \beta_{o})$ at time $t$, the object mesh is transformed as:
\begin{equation}\label{eq:object-based-v}
\begin{split}
\hat{\mathbf{V}}(\bTheta_{o}^t) &= \{ R_{\bphi_{o}} \left( \Psi_{\btheta_{o}^t}(\beta_{o}\mathbf{v}_{i}) \right)+ \boldsymbol{x}_o \}, 
\end{split}
\end{equation}
where $\beta_{o} \in \mathbb{R}^3$ is scaling factor, $R_{\bphi_{o}}$ is the global rotation matrix, and $\boldsymbol{x}_o$ is the global translation. $\Psi_{\btheta_{o}^t}$ represents the part mobility matrix which is dependent on the selected parametric model and part motion value at time $t$. $\Psi_{\btheta_{o}^t}$ is the identity matrix for $\mathbf{v}_i \in \mathbf{V}_{base}$ since the object base is static with no motion.  

For revolute motion, $\Psi_{\btheta_{o}^t}$ is a rotation matrix defined by the rotation axis of corresponding SAPIEN model together with the part rotation angle $\btheta_{o}^t$ at time $t$. Note that matrix $\Psi_{\btheta_{o}^t}$ is fully differentiable w.r.t value $\btheta_{o}^t$. For prismatic-joint, we follow a similar approach and define $\Psi_{\btheta_{o}^t}$ to be a transformation matrix correspond to the given translation axis and the offset value $\btheta_{o}^t$. In this way, gradient can backprop from object mesh vertices back to the articulated object parameters.

\section{Objective Function \label{sec:final_loss}}
To estimate the pose, part motion, and shape of the articulated object, we minimize an objective function that is the sum of four error terms: object mask term $E_{\text{m}}$, human fitting term $E_{\text{h}}$, human-object interaction term $E_{\text{hoi}}$, and a regularization term $E_{\text{r}}$. The overall objective $E(\bTheta_{o}^t, \boldsymbol{x}_h^t, R_c) =$ 
\begin{equation}\label{eq:overall-obj-loss} 
\begin{split}
\lambda_{\text{m}} E_{\text{om}}(\bTheta_{o}^t) + \lambda_{\text{h}} E_{\text{h}}(\boldsymbol{x}_h^t; H_{\text{est}}^t) +\\ \lambda_{\text{hoi}} E_{\text{hoi}} (\bTheta_{o}^t, R_c; \boldsymbol{x}_h^t, H_{\text{est}}^t) + \lambda_{\text{r}} E_{\text{r}}(\bTheta_{o}^t).
\end{split}
\end{equation}
Here $\lambda_{\text{m}}, \lambda_{\text{h}}, \lambda_{\text{hoi}}, \lambda_{\text{r}}$ are the hyper-parameters, object parameters $\bTheta_{o}^t = ( \boldsymbol{x}_o, \bphi_{o}, \btheta_{o}^t, \beta_{o})$, $H_{\text{est}}^t$ is the estimated orthographic human mesh from~\cite{joo2020exemplar}, $\boldsymbol{x}_h^t$ are the translation parameters that transforms $H_{\text{est}}^t$ to the same 3D space as the object, and finally $R_c$ is the rotation matrix used for human-object contact curve matching. Next, we will describe those error terms in more details.

\subsection{Object Mask Error \label{sec:obj_only}}
The object mask term penalizes inconsistency between the projected 3D object mask and the annotated object mask $M(t)$ estimated from PointRend~\cite{kirillov2020pointrend}. It is based on a differentiable renderer~\cite{liu2019soft} implemented in PyTorch3D~\cite{ravi2020pytorch3d}. Their method provides fully differentiable projection $P_{\text{soft}}$ from 3D mesh vertices to 2D image mask. Here, we set up $P_{\text{soft}}$ as perspective projection using our estimated focal length $K$. After applying global transformation and part motion, object meshes are passed through the differentiable render and we get the rendered 2D mask at every frame. The final object mask error is defined as: 
\begin{equation}\label{eq:object-based-loss}
\begin{split}
E_{\text{m}}(\bTheta_{o}^t) = \frac{1}{N} \sum_{t=1}^{N} ||P_{\text{soft}} \left( \hat{\mathbf{V}}(\bTheta_{o}^t); \ K \right) - M(t)||^2,\\
\end{split}
\end{equation}
where $\hat{\mathbf{V}}(\bTheta_{o}^t)$ is defined as in Eq.~\ref{eq:object-based-v}.

\subsection{Human Fitting Error \label{sec:h_fit}}
Human fitting error measures the difference between projected human mesh vertex and the estimated 2D vertex locations $p$ (also from EFT~\cite{joo2020exemplar}) in the same frame.  Minimizing the human fitting term $E_{\text{h}}$ will transform the estimated orthographic human body $H_{\text{est}}^t$ into the same 3D space as the object. We use a perspective projection $P_{\text{persp}}$ similar to $P_{\text{soft}}$ with the same focal length $K$ and scale the human mesh with a pre-determined scale $S_h$ so that the SMPL mesh height is equal to the height of the volunteer appearing in the video (170 cm for all videos). We then add 3D translation offset $\boldsymbol{x}_h^t$ to all the vertices at each time frame. The human fitting error is defined as $E_{\text{h}}(\boldsymbol{x}_h^t; H_{\text{est}}^t)=$
\begin{equation}\label{eq:h-fit-loss} 
\begin{split}
\frac{1}{N} \sum_{t=1}^{N} ||P_{\text{persp}} \left( S_h H_{\text{est}}^t +\boldsymbol{x}_h^t; K \right) - p(t)||^2.\\
\end{split}
\end{equation}
Note that we allow per-frame 3D translation to be applied to the human mesh so that human can move around in the video.

\subsection{Human-object Interaction Error \label{sec:obj_pp_joint}}
After inserting human into the same 3D world coordinate as the object, we now describe our human-object interaction term. This consists of two separate parts, orientation error and contact error. We will now describe each one in details.

\paragraph{Orientation error}
Our first human-object interaction term is the orientation error. During interaction, the human often faces towards the interacted object. Motivated by this observation, we estimate the human facing direction from SMPL and encourage object front direction to be the opposite. We define human facing direction $D_f(t)$ as the cross product of two vectors: SMPL left shoulder to SMPL right hip, and SMPL right shoulder to SMPL left hip. Object front direction is the vector pointing from the object center towards the direction where human interaction most frequently occurs. For our post-processed data, this direction is $+z$ in the local object coordinate before applying global rotation.

In addition to human facing direction, we also want to align human and object so that their inferred ground normal directions are parallel. We define ground normal direction $D_g(t)$ from human as the cross product of two vectors: left SMPL feet ankle to right SMPL feet toe, and right SMPL feet toe to left SMPL feet ankle. The object ground normal direction is always pointing towards $+y$ direction in the local object coordinate. The overall human orientation error is defined as $E_{\text{orientation}}(\bTheta_{o}^t; \boldsymbol{x}_h^t, H_{\text{est}}^t)=$
\begin{equation}\label{eq:hoi-orien-loss} 
\begin{split}
\sum_{t=t_\text{start}}^{t_\text{end}}( cos(D_f(t), R_{\bphi_{o}} \vec{-z}) +\cos(D_g(t), R_{\bphi_{o}} \vec{y})).
\end{split}
\end{equation}
where $R_{\bphi_{o}}$ is the object global rotation matrix, $t_\text{start}, t_\text{end}$ are the start and end time of the interaction where the object part is moving.

\paragraph{Contact error}
Our second human-object interaction term is the contact error. We make the assumption that during human-object interaction, the object motions tends to follow the human hand. This allows us to use the 3D hand location from the human mesh to constrain the 3D location of moving part. 

One issue that arises is that it is difficult to find the exact contact vertex on the object when human is touching it. During optimization, object orientation and location in general do not closely match the ground-truth. Matching the 3D human-object contact location introduce noise and might even lead to decreased performance. To solve this issue, we introduce a pose-invariant contact term that can tolerate such noise by only matching the 3D shape of the human-object contact curves. We select key vertex locations on the 3D CAD models which are most representative of the shape of the moving part during interaction. During optimization, we match against both the left and right hands with the selected object contact vertex and report the final result with the lowest optimization cost. 

Assume our selected object contact vertex is $v_{c}$. We then have the same vertex in 3D camera coordinate (world coordinate) $\hat{\mathbf{v}}_{c}(t)$ for each frame. Similarly, we can find the hand contact vertex $\hat{\mathbf{h}}_{c}(t)$. During our experiments we find that it is sufficient to use the vertex located at the center of the palm. To make the curve matching pose-invariant and only focus on shape, we allow rigid transformation of the contact curve during matching as this does not change the shape of the curve. The pose-invariant human contact error is defined as $E_{\text{contact}}(\bTheta_{o}^t, R_c; \boldsymbol{x}_h^t, H_{\text{est}}^t)=$ 
\begin{equation}\label{eq:hoi-contact-loss} 
\begin{split} 
\sum_{t=t_\text{start}}^{t_\text{end}} ( R_c  \hat{\mathbf{v}}_{c}(t) + t_c - \hat{\mathbf{h}}_{c}(t) )^2.
\end{split}
\end{equation}

Here $R_c$ is the curve rotation and $t_c$ is the curve translation. In practice, we can directly compute $t_c$ as the offset between the first frames. Note that the rigid transformation is not frame-dependent. 

\subsection{Regularization\label{sec:constraint}}
We use several regularization to avoid unrealistic results. The final regularization term $E_{\text{r}}$ includes 1) mask center error for pulling object mask closer to the annotated mask, 2) part motion error for penalizing out of limit motions, 3) depth error for making human and object depth closer during interaction, 4) global orientation error that penalizes large object roll or large negative pitch values (camera located beneath object), and finally 5) smoothing error that avoids sudden part motions. All these terms are formulated using the L2 loss.

\section{Optimization \label{sec:optim}}
We perform gradient-based optimization with the objective goal: $\text{arg min} \ E(\bTheta_{o}^t, \boldsymbol{x}_h^t, R_c)$ as defined in Eq.~\ref{eq:overall-obj-loss}. Without assuming more information about the contact state, we optimize all possible combinations consisting of left/right hand, all object models from the corresponding category, and all possible object contact vertex locations. After optimization, the combination with the lowest $E(\bTheta_{o}^t, \boldsymbol{x}_h^t, R_c)$ is chosen as our prediction. Adam~\cite{kingma2014adam} is used in all optimizations. The initial learning rate is set to $0.05$ and reduced to $0.005$ in the last 50 iterations. $\beta_{1}, \beta_{2}$ in Adam are set to $0.9, 0.999$ respectively. We optimize for a total of 200 iterations.

\vspace{20pt}

\section{Experimental Results}
We first discuss evaluation metrics for benchmarking our dataset in Section~\ref{sec:eva_met}. Our optimization results are available in Section~\ref{sec:quan}. We conduct ablation studies in Section~\ref{sec:abl}.  Finally we provide qualitative results in Section~\ref{sec:qual}. 

\subsection{Evaluation Metrics \label{sec:eva_met}}
In our experiments, we use five different metrics for evaluating the pose, part motion and shape of the reconstructed object. We measure the difference in object orientation as the relative angle (in degrees) between estimated rotation matrix and the ground-truth rotation matrix (computed from Euler angles) in the SO(3) space. Object location error is measured as the 3D distance between estimated and ground-truth object base center (in cm). We calculate part motion error using absolute difference in degrees for revolute motion and absolute difference in length (cm) for prismatic motion. We also measure the averaged object dimension difference (in cm). Results in Section~\ref{sec:quan} are optimized with the ground truth CAD model being provided as part of the input. In the ablation studies, we also conduct studies where different models are considered and find the best matching one. In this case, we also report the CAD model correctness accuracy. 

\begin{table}[!htbp]
\resizebox{0.475\textwidth}{!}{
\begin{tabular}{@{}lcccc@{}}
\toprule
\multicolumn{1}{c}{} & \begin{tabular}[c]{@{}c@{}}orientation \\ (degree)\end{tabular} & \begin{tabular}[c]{@{}c@{}}location\\ (cm)\end{tabular} & \begin{tabular}[c]{@{}c@{}}part motion \\ (degree/cm)\end{tabular} & \begin{tabular}[c]{@{}c@{}}dimension\\ (cm)\end{tabular} \\ \midrule
dishwasher           & 10.134                                                          & 20.941                                                  & 12.158                                                             & 12.287                                                   \\
washingmachine       & 19.791                                                          & 28.673                                                  & 19.634                                                             & 12.243                                                   \\
refrigerator         & 11.301                                                          & 31.390                                                  & 13.343                                                             & 11.013                                                   \\
microwave            & 22.504                                                          & 18.901                                                  & 14.560                                                             & 10.991                                                   \\
laptop               & 29.379                                                          & 29.535                                                  & 23.530                                                             & 8.469                                                    \\
trashcan             & 30.738                                                          & 83.530                                                  & 14.890                                                             & 20.969                                                   \\
oven                 & 7.059                                                           & 14.644                                                  & 14.273                                                             & 12.788                                                   \\
storage (revolute)   & 12.036                                                          & 16.542                                                  & 17.486                                                             & 6.860                                                    \\
storage (prismatic)  & 19.252                                                          & 23.785                                                  & 3.402                                                              & 14.210                                                   \\
average  & 18.021                                                          & 29.782                                                  & 16.234                                                              & 13.729                                                   \\
\bottomrule
\end{tabular}}
\caption{Object state estimation error for full human-object interaction objective. Category-level error is reported using averaged result over all frames within one category. Last row provides average over all categories. Averaged part motion excludes storage (prismatic) due to its values been in cm rather than degrees. }
\label{tab:ours_result}
\end{table}

\subsection{Quantitative Evaluation \label{sec:quan}}
In this section, we provide quantitative results for our optimization-based method. We simplify the problem by using the ground truth CAD model. Note that we still need to estimate the object dimension. We use the ground-truth object masks (rendered from ground-truth object pose and motion provided in the dataset). Results for more challenging cases are examined in the ablation studies. 

Table~\ref{tab:ours_result} summarizes the results on all eight categories. We separately report storage-furniture results based on whether the moving part exhibits revolute or prismatic motion. We make several observations from the results. First, we see that part motion error is low in cases where small motion changes lead to large differences in the object mask. The part motion error for trashcans is the largest because the trashcan lid is very small and can not be easily reflected in our differentiable object mask term. Trashcans also have large location error because some of the captured trashcan data can not be correctly approximated well by the CAD models we use. Human pose estimation from  Joo et al.~\cite{joo2020exemplar} can also be noisy due to occlusions (e.g. laptop data where the laptop occludes most of the human body). The dimension error for large objects like refrigerators can also be very small. This indicates that it is not the case that large objects will always have larger size errors.

\subsection{Ablation Studies \label{sec:abl}}

\paragraph{Analysis of HOI term}

\begin{figure}
\includegraphics[width=0.5\textwidth]{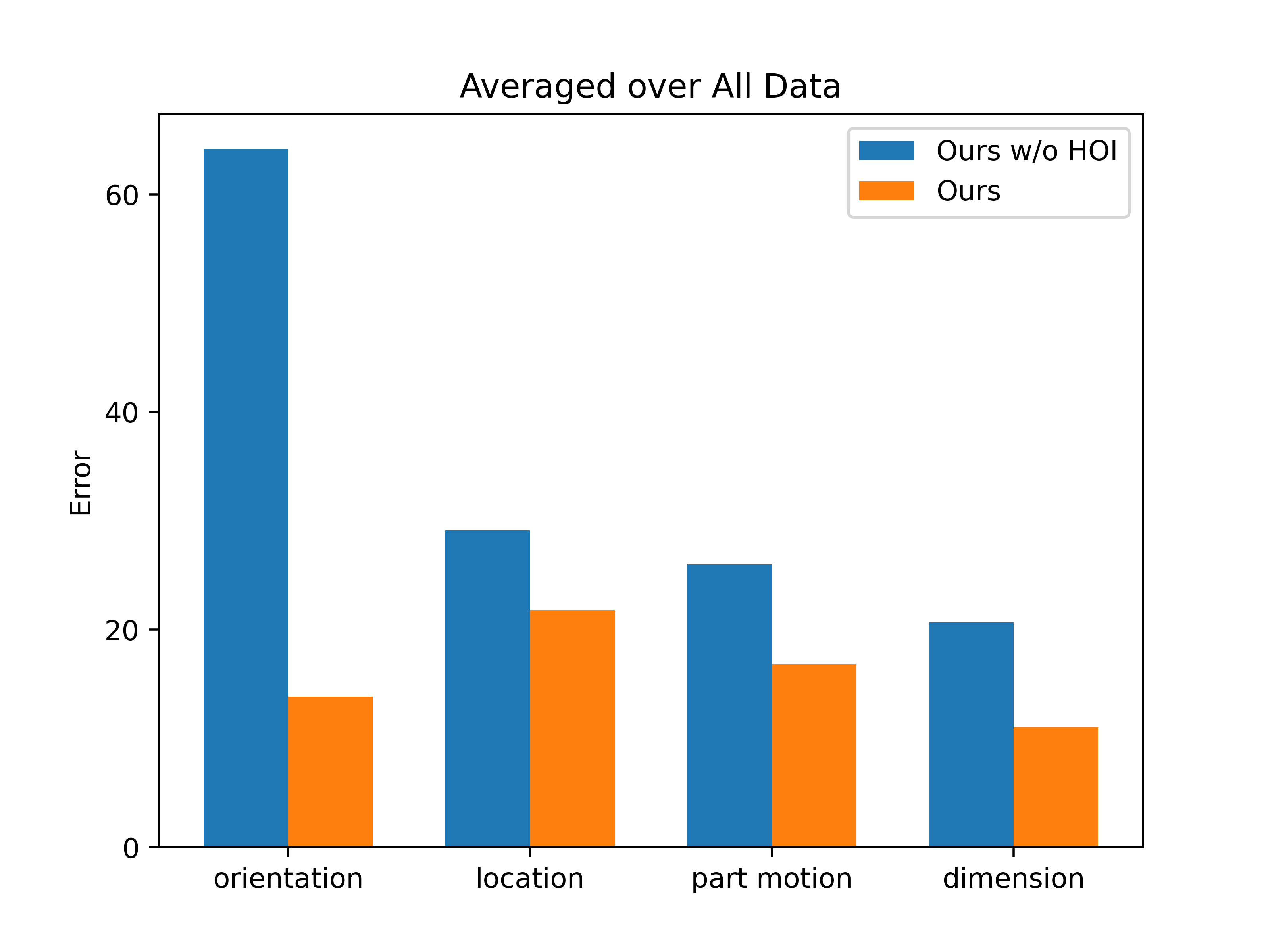}
\caption{Comparison of full human-object interaction objective against object-only objective in terms of overall articulating part estimation error metrics for dishwasher, oven, storage (revolute), and washingmachine categories. Error is in degrees for orientation and part motion, and in cm for location and dimension. Lower is better. We see that removing the HOI terms significantly decreases performance.}
\label{fig:compare1}
\end{figure}

To understand the effect of human-object relations, we compare results when optimized without human-object interaction error. The regularization term $E_{\text{r}}$ is still used except for the human-object depth error. Results are averaged over all four best-performing categories (dishwasher, oven, storage revolute, and washingmachine). Figure~\ref{fig:compare1} summarizes this comparison. We observe that optimized results deteriorate significantly without human-object interaction terms. The most decrease comes from global object orientation. This is expected since the 2D object mask does not contain much useful information to constrain the object pose.

\paragraph{Analysis of CAD model} Previously, we assumed that the ground truth CAD model is given. In practice, we do not have access to this information. Thus it is necessary to optimize the selection of CAD model together with the rest of the parameters. To this end, we optimize over all CAD models in the corresponding category and keep the setting with the lowest objective cost. We test this on the dishwasher and storage (revolute joint) categories. The CAD model correctness accuracy is $90.91\%$ and $71.43\%$ for the two categories, respectively. When the retrieved model is incorrect, other errors will be large. This demonstrates the need for classification or retrieval approaches that can find the closest-matching CAD models.

\subsection{Qualitative Evaluation \label{sec:qual}}
\begin{figure}
\includegraphics[width=0.9\linewidth]{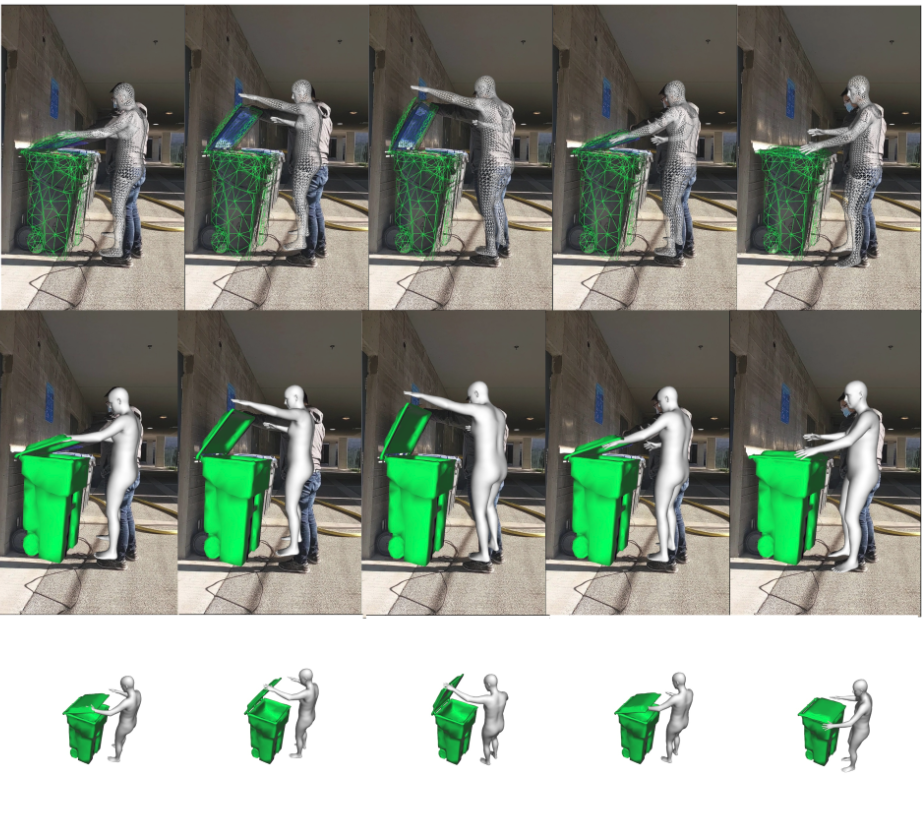}
\includegraphics[width=0.9\linewidth]{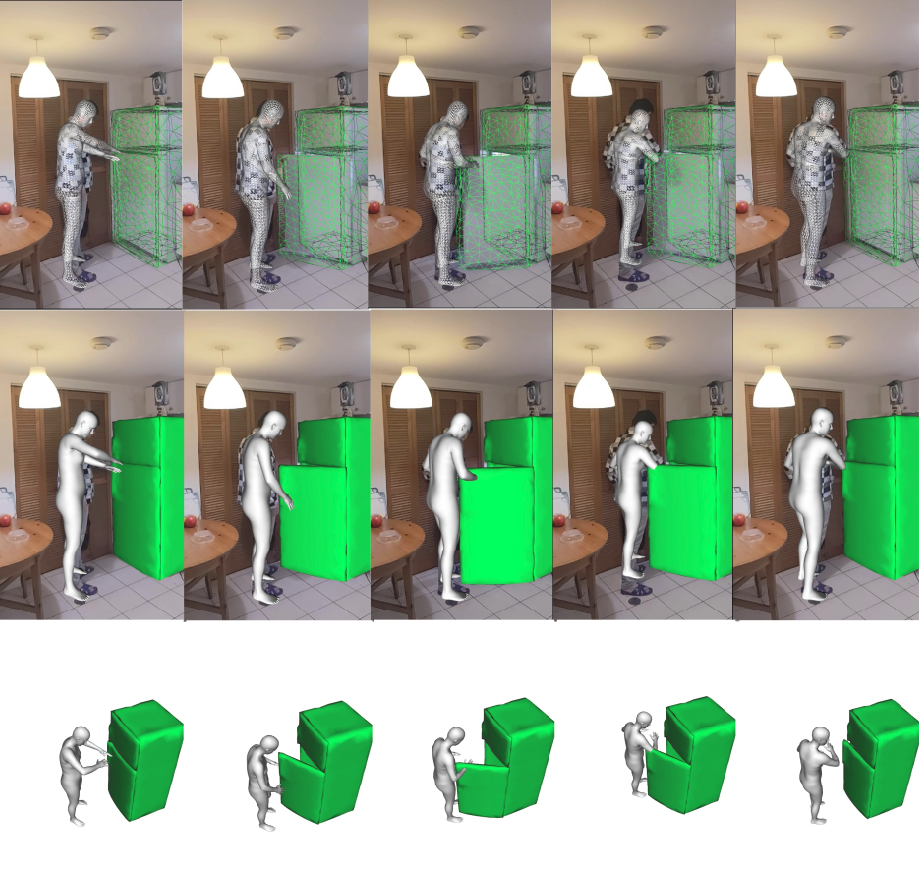}
\includegraphics[width=0.9\linewidth]{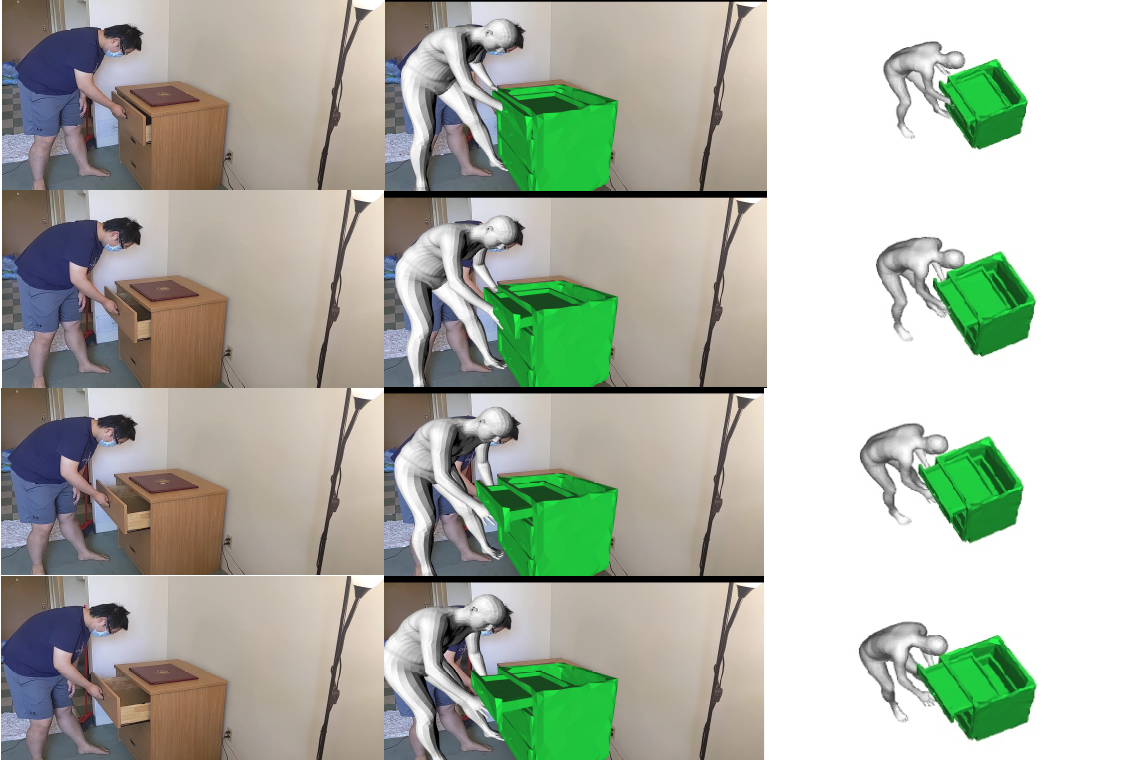}
\caption{Results using the full human-object interaction objective optimization. The human and object layout is visually correct.}
\label{fig:qual1}
\end{figure}

Figure~\ref{fig:qual1} illustrates several example outputs after our optimization. Example videos of the results are available \href{https://www.youtube.com/watch?v=LhZ1TkRUznA}{here}. We observe that even though the human mesh is fixed in that we do not optimize its shape or joint angles, the human and object layout is visually plausible. The object dimensions ratio also look plausible and the object part moves accordingly as the human interacts with it.

\begin{figure}[!htbp]
\includegraphics[width=\linewidth]{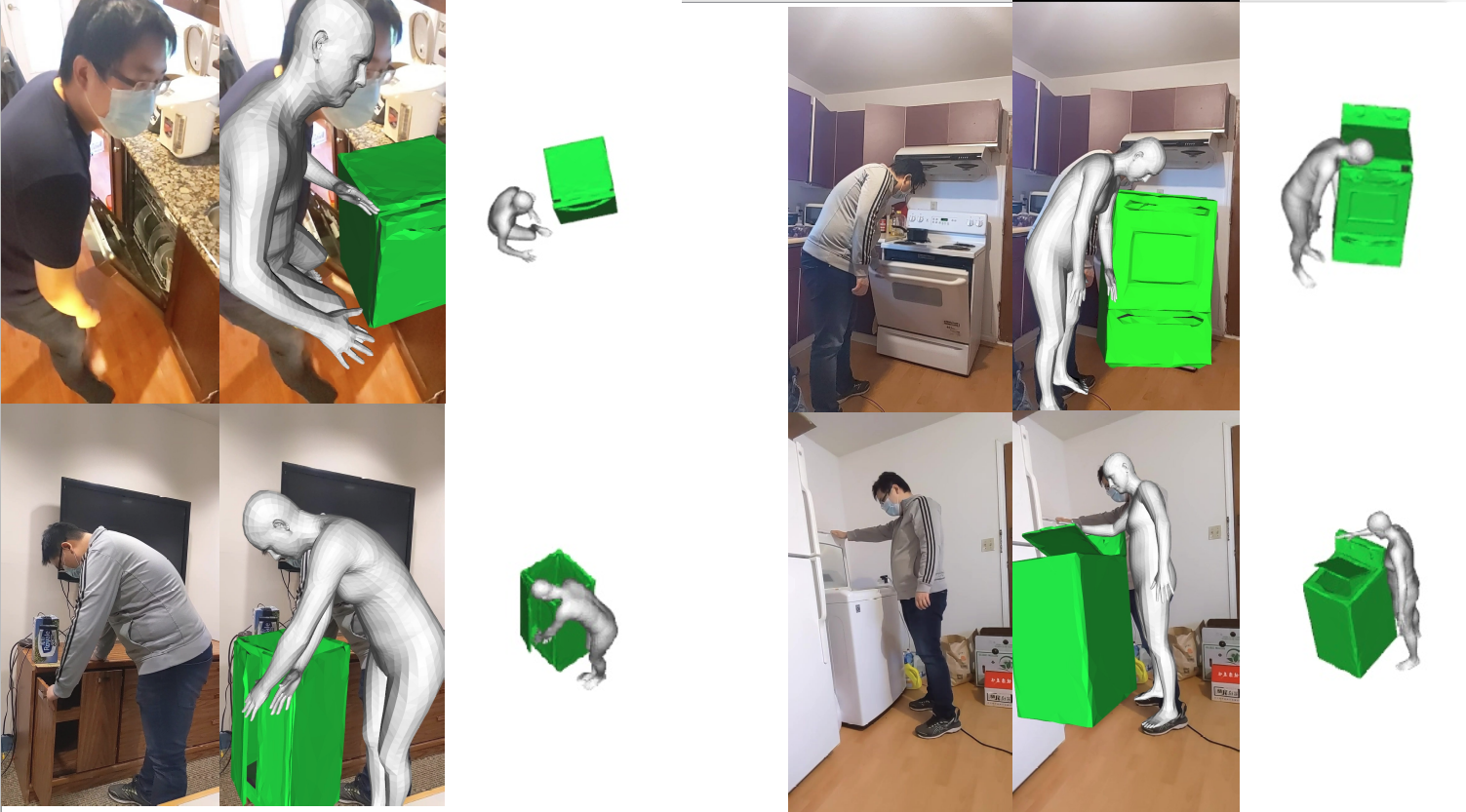}
\caption{Results without the HOI term. }
\label{fig:fail}
\end{figure}

We also visualize the optimization results without the HOI term. Results are shown in Figure~\ref{fig:fail}. Without human-object relations, we find that the object orientation is incorrect in many cases. To still match the object mask at each frame, the estimated part motion is rather inaccurate. This either leads to large sudden motions or no motion. The same goes for the object dimension where the height, width, and length ratio is incorrect (e.g. oven). 

\vspace{20pt}

\section{Conclusion}
We introduce the D3D-HOI Dataset with frame-level annotation for object shape, pose, and part motion. Our dataset provides a real-world 3D benchmark for object pose and motion estimation during human object interaction. It can also be extended for training generative model for human-object interaction snippets. We propose an optimization-based methods and show that modelling human-object relations can improve the estimated 3D object parameters. We hope this work will motivate further research in 3D human-object interactions.

\section{Acknowledgement}
We thank all five volunteers: Lelei Zhang, Yongsen Mao, Bingqing Yu, Hanxiao Jiang, and Sam Xu for helping with the data collection.

{\small
\bibliographystyle{plainnat}
\setlength{\bibsep}{0pt}
\bibliography{egbib}
}

\end{document}